\documentclass{article}

\usepackage{microtype}
\usepackage{graphicx}
\usepackage{amsmath}
\usepackage{booktabs} 
\usepackage{subfig}
\usepackage[algo2e]{algorithm2e}
\usepackage{amsfonts}
\usepackage{amssymb}
\usepackage{float}
\usepackage{fnpos} 

\usepackage{hyperref}

\usepackage[accepted]{icml2021}

\icmltitlerunning{Deep Active Learning Using Barlow Twins}

\begin{document}

\twocolumn[
\icmltitle{Deep Active Learning Using Barlow Twins}



\icmlsetsymbol{equal}{*}

\begin{icmlauthorlist}
\icmlauthor{Jaya Krishna Mandivarapu}{equal,to}
\icmlauthor{Blake Camp}{equal,to}
\icmlauthor{Rolando Estrada}{equal,to}
\end{icmlauthorlist}

\icmlaffiliation{to}{Department of Computer Science, Georgia State University, Atlanta, Unites States}

\icmlcorrespondingauthor{Jaya Krishna Mandivarapu}{jmandivarapu1@student.gsu.edu}
\icmlcorrespondingauthor{Blake Camp}{bcamp2@student.gsu.edu}
\icmlcorrespondingauthor{Rolando Estrada}{restrada1@gsu.edu}


\icmlkeywords{Machine Learning, ICML}

\vskip 0.3in
]



\printAffiliationsAndNotice{\icmlEqualContribution} 

\begin{abstract}
The generalisation performance of a convolutional neural networks (CNN) is majorly predisposed by the quantity, quality, and diversity of the training images. All the training data needs to be annotated in-hand before, in many real-world applications data is easy to acquire but expensive and time-consuming to label. The goal of the Active learning for the task is to draw most \textit{informative  samples} from the unlabeled pool which can used for training after annotation. With total different objective, self-supervised learning which have been gaining meteoric popularity by closing the gap in performance with supervised methods on large computer vision benchmarks. self-supervised learning (SSL) these days have shown to produce low-level representations that are invariant to distortions of the input sample and can encode invariance to artificially created distortions, e.g. rotation, solarization,cropping etc. self-supervised learning  (SSL) approaches rely on simpler and more scalable frameworks for learning. In this paper, we unify these two families of approaches from the angle of active learning using self-supervised learning mainfold and propose \textbf{D}eep \textbf{A}ctive \textbf{L}earning using \textbf{Barlow} \textbf{Twins } (DALBT), an active learning method for all the datasets using combination of classifier trained along with self-supervised loss framework of Barlow Twins to a setting where the model can encode the invariance of artificially created distortions, e.g. rotation, solarization,cropping etc.. We propose to use joint loss function which consist classifier loss and self-supervised loss borrowed from Barlow twins to jointly learn an encoder that produces representations invariant across such pairs. DALBT is a method that is simple, easy to implement and train, and of broad applicability. We carried out an extensive evaluation of our novel proposed method of active learning, achieving state-of-the-art results on MNIST, Fashion-MNSIT, CIFAR-10. Additionally, to show the robustness of the proposed model we also showed the results on where the unlabeled pool consists of a mixture of samples from multiple datasets, proposed model can successfully distinguish between samples from seen vs. unseen datasets.
\end{abstract}
\section{Introduction}Although deep neural networks (DNNs) demonstrated state of the art (SOTA) accuracy on several supervised learning tasks such as as classification \citep{he2016deep,krizhevsky2012imagenet}, object detection \citep{ren2015faster,redmon2016you}, and semantic segmentation. But most of the deep neural networks (DNNs)  require large set of labeled data to achieve this feet. The challenges of labeling huge datasets in real world setting are many: expensive, limited time available by domain business experts, long labeling time per for large-scale sample such as videos and time-series data, financial constraints, or to minimize the model’s carbon footprint. These all drawback does inherit the application of deep neural networks (DNNs) to more research areas and more organization.\\

In order to overcome the above drawbacks, Active Learning(AL) system try to select to most informative samples from the pool of unlabeled data points at each stage and send them for annotation to maximize the accuracy of the model. Active learning uses a fixed budget at each stage of learning  to select and label a subset of a data points from the unlabeled pool where budget($b$) refers to cost associated with annotation by oracle($\mathcal{O}$). The model will be trained on the current labeled pool along with the newly annotated data points. At the end of active learning process model’s performance would be nearly the same accuracy as model by utilized fraction of data when compared to the model trained on all the data. Active Learning(AL) also highlights the fact that there exists a non-linear relationship between the model’s performance
and the amount of training data used. There exists most representative subset of the unlabeled data and selecting those data points to label will provide most of the information needed to learn to solve a
task. In this case, we can achieve nearly the same performance by selecting that representative subset for annotation (and training on) only using data points from that representative subset samples, rather than the entire dataset.\\

In contrast self-supervised learning which learns useful information from the dataset without relying on human annotations. Most of the work in the field of self-supervised learning work on goal of leaning good low level representations of input data without access to data labels. With the current advances in the field of Self-supervised learning (SSL)which is rapidly closing the dap with supervised learning methods on large datasets and computer vision taks. Most of the methods in SSL work with the goal of learning representations that are invariant under different distortions such as random cropping, resizing, horizontal flipping, color jittering, converting to grayscale, Gaussian blurring, and solarization.(also referred to as ‘data augmentations’).\\

From the high level view  Active learning reduce the label-effort and Self-supervised learning aim to use the unlabeled data. With the goal of  merging self-supervised learning along with supervised learning we propose \textbf{D}eep \textbf{A}ctive \textbf{L}earning using \textbf{Barlow} \textbf{Twins } (DALBT). The proposed method aims to combine self-supervised learning along with supervised learning in each step. Our proposed work is different from the previous work in the field which achieved this by doing pre-training on all the entire/partial unlabeled dataset using self-supervised learning then use the pre-trained model for active learning training step using supervised learning loss. But this approach increases the overall training time as in some cases the overall size of the  unlabeled pool can be really big and this process is not feasible at times. In this paper we propose an Active learning system that utilizes both the supervise learning and unsupervised learning to achieve the goal of selecting the most informative samples from the unlabeled pool.\\

Our paper is organized as follows: In section 2 we describe the related work. Next, in section 3 we introduce
the proposed framework. Section 4 present the experimental setup and the evaluations on the datasets we used.
Finally, section 6 conclusion and discusses an interesting finding we observed in the proposed work.

\begin{figure*}[t]
  \centering
    \includegraphics[width=0.8\textwidth]{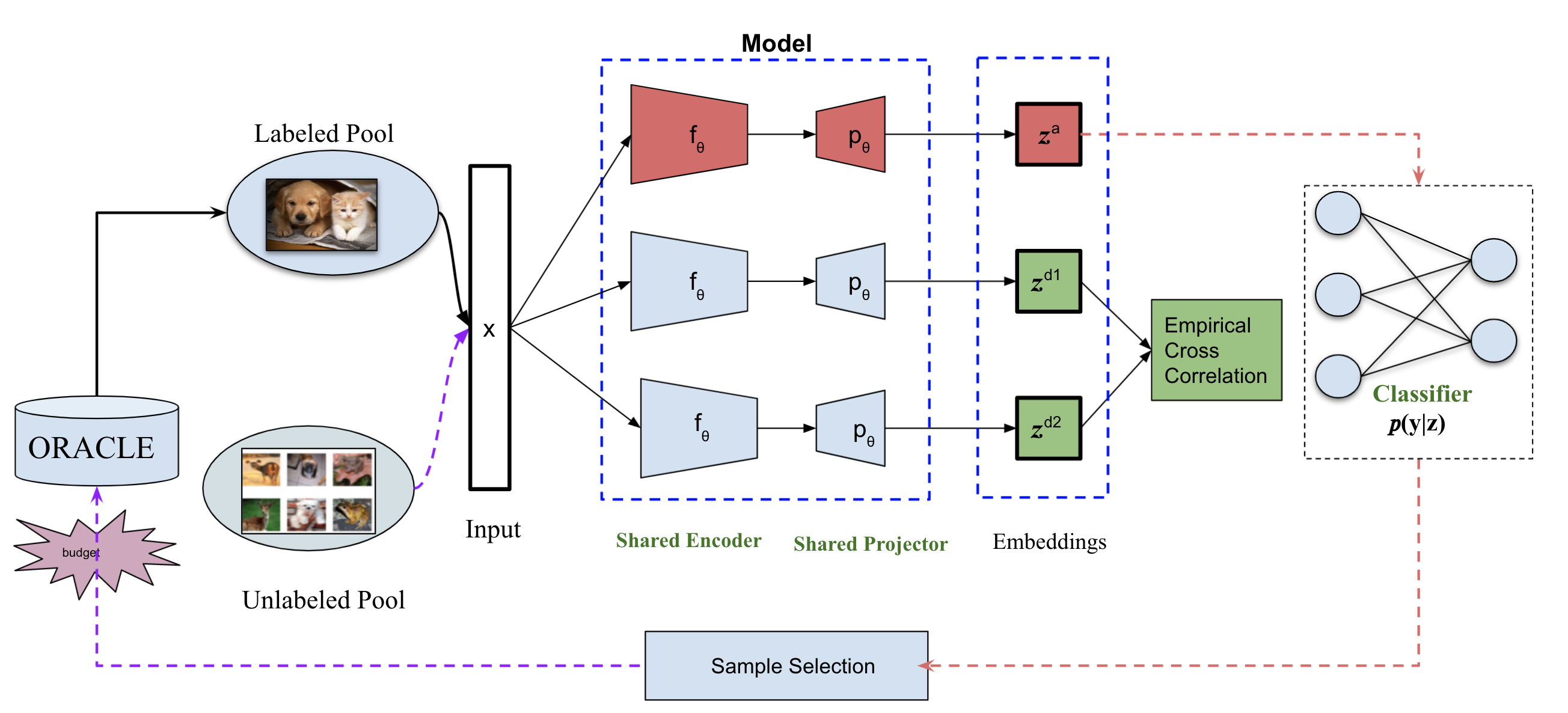}
  \caption{\textbf{Framework overview:} Our proposed active learning system based on BarlowTwins consists of encoeder($E$),Projector($P$),Classifer(C). Encoder takes two distorted versions of same input as its input. Output of encoder is fed into projector network which projects both the distorted versions into lower dimension. Classifier take latent vector($z$) produced by passing the actual input image without distortion through the encoder($E$). Overall system is trained using joint loss of cross-corelation loss and classification loss. Wiebull Sampling method to identify which samples from the unlabeled pool to label.}
  \label{fig:architecture}
\end{figure*}


\section{Related Work}
Here we detail previous work done in each of these directions.
\label{sec:relatedWork}
\label{sec:priorWork}
\subsection{Active Learning}
 Active learning methodologies were recently reviewed by Settles\citep{activeLearningLitSurvey}), discussed more here \citep{dasgupta2011two, hanneke2014theory} and it has been shown that there exists informative samples which contribute to performance than the other training samples. Thus, the overall
goal of active learning is to learn or use an acquisition function along with model to chooses the best data points for which
a label should be requested from a large unlabeled pool of
data. 

Existing Active learning approaches can be divide into Pool based methods or Query Synthesizing methods. Pool based methods tries to find the most informative samples from the unlabeled data using different sampling strategies which are more discussed in detailed in this section. Query Synthesizing methods \citep{AL_usingSampleSelectionConditionalGenerativeAdversarialNetwork,EMandPoolBasedALforTextClassification,DBLP:journals/corr/ZhuB17} use generative models to genearate the informative samples.\\

Active learning sampling strategies can be sub-divided into following catgeories a) \textbf{Uncertainity Sampling} This is one of the popular sampling methodology in which  where the model queries
data points about which it is most uncertain about. Recent research \citep{costEffectiveActiveLearningMelanoma,DBLP:journals/corr/WangZLZL17,powerOfEnsemblesForActiveLearning, lewis1994sequential, scheffer2001active} shows that uncertainity sampling approaches have proven effective in deep learning models such as CNNs. b) \textbf{Diversity sampling} This sampling method aims to choose samples which are more diversify from the existing labeled samples.  c) \textbf{Representative Sampling} This sampling method aims to choose the samples from the unlabeled pool which are representative of the whole dataset. There exists subfield in the research which uses combination of features from these three disjoint groups mentioned above to increase the performance of activelearing system.\\

\citep{schohn2000less} uses active learning to enhace the performance on document classification tasks using support vector machines (SVM) by  labeling examples that lie closest to
the SVM’s dividing hyperplane. The authors also proposed stopping heuristic for AL on when the model reach the peak generalization performance.\\

\citep{tong2001support} applied active learning to  Text Classification using SVM by implicitly projecting the training data into a different (often higher dimensional) feature space which is linearly separable. Then projects the query selection problem as version space optimization problem in which version space optimizates as quickly as possible still obeying the SVM constraints which is equivalent to finding informative samples quickly.

\citep{tur2005combining} combines active and semi-supervised learning methods int the domain of spoken language understanding.


\citep{wang2017uncertainty} combines uncertainty based active learning 
algorithm with diversity constraint by sparse selection in which sample selection is represented as a sparse modeling problem.\\
\citep{sener2017active} projected the problem of active learning as core-set selection problem in set of points are choosen such that selected points should be dissimilar both to each other and the labelled set, representative of the unlabelled set and  competitive for the remaining data points\\

\citep{zhu2009active} combined  uncertainty and
density (SUD) and density-based re-ranking to overcome the problem outlier selection problem present in Uncertainty sampling. By combining uncertainty along with density-based re-ranking which selects the samples which are most informative example in terms of uncertainty criterion, but also the most representative example in terms of density criterion.\\


Similar to core-set approach of selecting batch of images in pool-based setting \citep{geifman2017deep} selects the points for each class by using farthest-first(FF) traversal principle or famously known as Gonzalez algorithm \citep{gonzalez1985clustering}. FF principle states that traversal for a set of points can be constructed by selecting the first point x randomly then next point is selected which is farthest from previously selected point x by greedily choosing the point farthest away from any of the points already chosen.
Set of point obtained using the neural activation over a representation layer by forward passing all the unlabeled data. farthest-first(FF) traversal principle is similar to building long-tail wiebull distribution.\\

\citep{gissin2019discriminative} motivated by selecting sample for which the probability of distinguishing it unlabeled pool and labeled pool is the highest. 
Such that selecting such kind of the samples for labeling should be informative and helps in increasing the performace of the model.\\

\citep{beluch2018power} showed ensembles perform better and lead to
more calibrated predictive uncertainties which can be used for ActiveLearning Uncertanity strategy. Authors also showed that this method performs better than the Monte-Carlo Dropout and geometric approaches\\

\subsection{Self-Supervised Learning} In recent years, self-supervised learning has achieved comparable performance w.r.t to suprvised learning \citep{caron2020unsupervised,chen2020simple,grill2020bootstrap}. Most of the self-supervised learning methods work with a goal of achieving where representations are learned that
are invariant to distortions present in the input data. Distorted inputs are created using different data augmentation applied to input randomly. Different research methods try to achieve this goal using different approaches such as in SIMCLR \citep{chen2020simple} achieved this by creating ‘positive’ and ‘negative’ sample pairs from the input data and treating each pair differently in the loss function,  BarlowTwins \citep{zbontar2021barlow} achieves this using variance and invariance terms in which two distorted versions of single sample should produce sample low level represntaion which is achieved using custom loss function consists of variance and redundancy reduction term. \\

\citep{ash2019deep} proposed  Batch Active learning by Diverse Gradient Embeddings (BADGE) method in which d to incorporate both predictive uncertainty and sample
diversity into every selected batch. Authors achieved this by calculating the gradient embedding for hypothetical label and used Kmean++ seeding algorithm  to choose the batch to be labelled. 

Previous works in the field of researchers trying to merge active learning and self-supervised learning are in Graphical domain \citep{https://doi.org/10.48550/arxiv.2010.16091} applied self-supervised learning along with active learing to Graph Neural Networks by considering the information propa-gation scheme of GNN and selecting the central nodes from homophilous ego networks, \citep{Bengar_2021_ICCV} utlized autoencoder architecturem SSL technique SIMCLR to form postive and negative pairs. In NLP for text classification task \citep{https://doi.org/10.48550/arxiv.2010.09535} used self-supervised learning as a pre-training step for training the language model and the samples which the language model is uncertainare sent for labelling and for efficient fine-tuning. Similar approach of large-scale pseudo training data by randomly adding or deleting words from unlabeled data is followed by \citep{wang2021combining} for disfluency
detection heavily rely on human-annotated data for solving sentence classificationt task. \citep{Bengar_2021_ICCV} Model is trained on the entire dataset to get the frozen backbone. Now linear classifier or an SVM, decoder is fine-tuned on top of the features in supervised way, inference is run on the entire unlabeled data and top-k samples are collected via acquisition function. In medical domain \citep{9361645} collect the salency maps of medical images and project it as self-supervised learning problem where the autoencoder reconstructs the saliency maps of medical images followed by clustering the latent space to collect the top-k and Query labels of the most representative sample per cluster. Our work is particulary different from the other work in the field as all the previous work concentrates on high amount of pre-training on all the data which is not feaisable as the unlabeled pool size is pretty high which add hughe overhead training time, creating "positive" and  "negative" pairs for training is not feaisable when the overall dataset size is pretty large. \citep{mandivarapu2020deep} merged the fields of active learning and open-set recognition in which model is trained using information bottleneck loss along with wiebull long tail distribution to find the outlier per class and achieved the state of the art in the field of active learning. \\

This proposed approach explores a active learning method including self-supervised learing which can be used for informative sample selection for labelling, in which the
wiebull sampling was used as acquistion function. This work proposes to address all of above mentitone issues with a single approach, driven by a distinct business need. 


\section{Methodology} \label{sec:methodology}
In this section, we briefly review the setup of the pool based active learning for computer vision classification tasks. We also discuss about used self supervised learning approach barlow twins and the intuitions behind using it. We then describe our proposed approach Deep Active Learning using Barlow Twins (DALBT). Throughout the sections, we refer to the model being trained as $f$ and denote its corresponding weights/parameters
$\theta$. Given an unlabeled pool($\mathcal{U}$) of examples X without label, in each sampling iteration, our sampling method selects a diverse set of examples on which the model is least confident and useful for training at next iteration of active learning.

\subsection{Problem Definition}
Formally, pool based active learning problem is denoted as $P=$ $( \mathcal{D}_{\text {train}}, \mathcal{D}_{\text {test}})$; $\mathcal{D}_{\text {train}}$ is  the training set from where initial pool of samples are taken. $\mathcal{D}_{\text {train}}$ can be sub divided into $( \mathcal{D}_{\text {L}}, \mathcal{D}_{\text {U}})$; $\mathcal{D}_{\text {L}}$ is  the labeled pool where each sample consist of pair of input and label denoted by ($( x_{\text {L}}, y_{\text {L}})$). $\mathcal{D}_{\text {U}}$ denotes a much larger pool
of samples ($( x_{\text {L}}$ ) which are not yet labeled. The goal of the active learning model is to train on labeled pool($( x_{\text {L}}$) and used it along with sampling method to iteratively querying the most label-efficient samples present in the ublabeled pool $\mathcal{D}_{\text {U}}$ to be annotated
by the oracle such that the expected loss is minimized by a fixed sampling budget(b). b is the total no of most informative samples that can be selected from the unlabeled pool at each stage of active learning setup. These selected b sampled will be sent to oracle for annotation.  We denote the state of a subset at a given timestep as $\mathcal{L}^t$ and $\mathcal{U}^t$, respectively, for $t \in\{0,1, \ldots\} $ where $t$ indicates the current stage of active learning stages. \\

In standard pool based active learning setup, we train model our active learning model($f$) with parameters $\theta$ on the initial labeled pool ($\mathcal{L}^0$) at stage t=0. After the initial stage t=0, $b$ datapoints are sampled from the unlabeled pool using some predefined sampling method (eg: uncertainty measure, confidence estimate ..etc ). These selected $b$ data points will be removed from unlabeled pool ($\mathcal{U}^0$) and sent to oracle ($\mathcal{O}$) for annotation. These annotated datapoints are then added to labeled pool ($\mathcal{L}^0$) which now becomes labeled pool (c) and unlabeled pool($\mathcal{U}^0$)  becomes ($\mathcal{U}^1$). Now the model again will be trained on new labeled pool ($\mathcal{L}^1$) at next stage t=1. In the current experimental setup we consider two scenarios where unlabeled pool($\mathcal{U}$) contains samples from the same datasets and mixture of multiple datasets.

\subsection{Active Learning System}
With the goal of active learning using self-supervised learning. Our system consists of and encoder($E$) followed by a projector ($P$), followed by a classifier($C$) as shown in the Fig \ref{fig:architecture}. Our goal is to learn an encoder $f_{\theta}$ can encode the invariance of artificially created distortions, e.g. rotation, solarization,cropping etc. The proposed model $f_{\theta}$ takes as input two distorted versions of the  vector $x \in \mathbb{R}^{D}$ and outputs a corresponding reduced vector $z=f_{\theta}(x) \in \mathbb{R}^{d}$, with $d<<D$. Without loss of generality, we define the encoder to be a neural network with learnable parameters $\theta$. Let $\mathcal{L}^0$  be a initial labeled pool training set of datapoints in $\mathbb{R}^{D}$, the $D$-dimensional input space. Let $x \in \mathbb{R}^{D}$ be a vector from $\mathcal{X}$ .

\subsubsection{Barlow Twins} 
As mentioned in the Section Introduction we have used Barlowtwins \citep{zbontar2021barlow} for finding the low-level representation of our inputs. In this we explain about barlow twins in more detailed fashion. Barlow twins networks consists of encoder($E$) appended with projector network (p) as shown in Fig \ref{fig:architecture} excluding the classifier. For simplicity of explanation let's consider the case where the batchsize is 1. For each input image two distorted versions are produced using different types of random data augmentations applied during the training. Lets consider $d^1$ and $d^2$ as two distorted version of same input image $x$. These two distorted inputs are then fed into encoder($E$) followed by a projector network ($p$) both with trainable parameters. The model then produces two output low level representation of the same input image but one each for each distorted version. Lets say $z^{d1}$ and $z^{d2}$ as two low-level representations of  $d^1$ and $d^2$.\\

Barlowtwins uses unique loss function as mentioned in the paper "" which is different from other SSL methods as shown below 

\begin{equation}
\mathcal{L}_{\mathcal{B} \mathcal{T}}  { \triangleq} \underbrace{\sum_{i}\left(1-\mathcal{C}_{i i}\right)^{2}}_{\text {invariance term }}+\lambda \underbrace{\sum_{i} \sum_{j \neq i} \mathcal{C}_{i j}^{2}}_{\text {redundancy reduction term }}
\end{equation}\\

where $\mathcal{C}$ indicates the cross-corelation matrix computed between the two identical networks which is given below and $\lambda$ is hyper-parameter for defining the importance between the first and second terms of the loss.

$$
\mathcal{C}_{i j} \triangleq \frac{\sum_{b} z_{b, i}^{d^{1}} z_{b, j}^{d^{2}}}{\sqrt{\sum_{b}\left(z_{b, i}^{d^{1}}\right)^{2}} \sqrt{\sum_{b}\left(z_{b, j}^{d^{2}}\right)^{2}}}
$$
where b indexes batch samples and i, j index the vector dimension of the networks’ outputs. C is a square matrix with size the dimensionality of the network’s output with range of values from -1 to 1 where -1 indicated no-corelation between the $z_{b, i}^{d^{1}}$ and $z_{b, i}^{d^{2}}$, where 1 indicated perfect corelation between  $z_{b, i}^{d^{1}}$ and $z_{b, i}^{d^{2}}$.\\

\subsubsection{Active Learning using Barlow Twins}
With the intention of merging both the self supervised learning method Barlow twins and active learning we proposed new changes to the existing barlow twins architecture as show in Fig \ref{fig:architecture} and explained further. We appended additional classifier($C$) to existing model of encoder($E$),projector($P$). Overall system is trained using the modified innovative loss as shown below

\begin{equation}
\mathcal{L}_{\mathcal{B} \mathcal{T}}  { \triangleq}   \underbrace{ \log p_{\boldsymbol{\xi}}(\boldsymbol{y} | \boldsymbol{z})}_{\text{Classifer term}} + \gamma*( \underbrace{\sum_{i}\left(1-\mathcal{C}_{i i}\right)^{2}}_{\text {invariance term }}+\underbrace{\lambda \sum_{i} \sum_{j \neq i} \mathcal{C}_{i j}^{2}}_{\text {redundancy reduction term}}) 
\label{combineloss}
\end{equation}

where $\gamma$ indicates the amount of importance given to the barlow twins loss and first term indicates the classifier loss. The overall system is optimized using the joint loss as shown in Eq \ref{combineloss}. As you can see that the input to the classifier is the latent vector produced by actualling passing the input image without distortions through the model($f$).

\subsection{Sampling technique}
With an aim to select $b$ most informative data points from the unlabeled pool along with the trained model.

In depth the objective of the loss function in the Eq \ref{combineloss} is finidng the low-level representations that captures as much information as possible about the inputs while being least informative about the distortion applied to the input. We used wiebull sampling technique proposed by \citep{weibull1951statistical} . Using the wiebull sampling method can be used to quantify weather a sample is an outlier or not. In our case if the latent representation is very different from the labeled pool latent representions it is considered as an outlier. Usage of long-tail distribution for finding the informative samples in the field of active learning is shown by \citep{gonzalez1985clustering} and  \citep{mandivarapu2020deep}. Firstly collect all the latent vectors of images which are classified correctly by the model at any stage of active learning. These latent vectors are sub-divided into the respestive clusters depending on their class label. Now mean of each cluster is calculated and distance between mean of each class to rest of the points was calculated. Wiebull distribution is modeled using these distances for each class cluster. Finally any new images with out label will be pass through the wiebull model to check the percentage by which this image sample is considered as an outlier for all images and top such images are collected for labeling or for getting annontated by the oracle.

\begin{algorithm}
  \SetKwData{Left}{left} \SetKwData{This}{this} \SetKwData{Up}{up}
  \SetKwFunction{Union}{Union} \SetKwFunction{FindCompress}{FindCompress}
  \SetKwInOut{Input}{input} \SetKwInOut{Output}{output} \SetKwInOut{Parameter}{}
  \textbf{Require}: Unlabeled pool $\mathcal{U}^0$, labeled pool $\mathcal{L}^0$,number of labeling iterations $t$, initialize $b$ (budget)\\
  \textbf{Require:} Active Learning Model( $f_\theta$), Optimizer\\
  \For{$k=1$ \KwTo $t$}
    {   
Train $f_\theta$ on Labeled Pool $(\mathcal{L}^{k})$\\
Z $\leftarrow$ Collect the latent vectors of all correctly classified samples in  Labeled Pool $(\mathcal{L}^{t})$\\
$\mathcal{C}_{{i}} \leftarrow$ Mapping of $Z_{i}$ onto separate cluster per class $\mathcal{C}_{i}$\\
$\mathcal{d}_{{i}} \leftarrow$ Calculate the distance of each point to its cluster $\mathcal{C}_{i}$\\
$\mathcal{W}_{{i}} \leftarrow$ Map the distances by fitting them to a wiebull model\\
 Z $\leftarrow$ Collect the latent vectors of all the samples  Unlabeled Pool $(\mathcal{U}^{k})$\\
 \For{$Z=1$ \KwTo $n$}
    {   
    Collect the outlier probability score using previously fitted wiebull model.
    }
    
    Request labels for top $b$ samples\\
    $L^{k+1} \leftarrow L^{k} \cup b$ samples. \\
    
    Train $f$ on  $L^{k+1}$.

    }
  \caption{Active Learning}\label{alg:csm_algo}
\end{algorithm}


\section{Experimental Results}
\label{sec:experiments}
We performed experiments on \textcolor{black}{four} image classification datasets: MNIST \citep{lecun2015deep}, CIFAR-10 \citep{krizhevsky2009learning}, and FashionMNIST \citep{xiao2017fashion}---following the methodology defined in Sec.~\ref{sec:methodology}. Below, we first present our implementation details, then discuss our results.

\subsection{Implementation Details} \label{Implementation}

\textcolor{black}{
\textbf{Hardware:} We carried out our experiments on a Dell Precision 7920R server with two Intel Xeon Silver 4110 CPUs, two GeForce GTX 1080 Ti graphics cards, and 128 GBs of RAM.}

\begin{figure}[h]
  \centering
  \includegraphics[width=8cm]{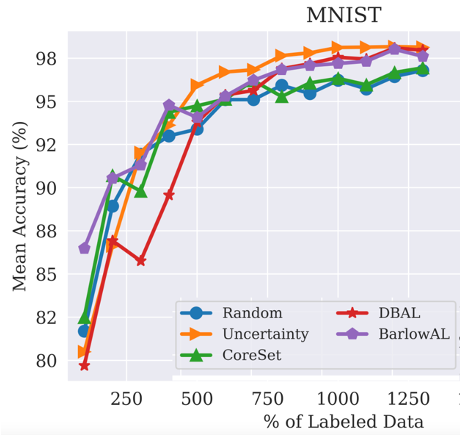}
  \caption{Robustness of our approach on MNIST classification . Our results confirm that our approach significantly outperforms this baseline.}
  \label{fig:mnist}
\end{figure}

\textbf{Dataset sizes and budgets:} \textcolor{black}{As previously explained in methodology section, \textit{budget} refers to the number of samples labeled by the oracle in each round of active learning. Budget of the each experiments is shown in the legend of the each result. MNIST dataset consists of 50,000 images as part of the  training set out of which is sub-divided into 100 images for the initial labeled pool, 5000 images  as a validation set, and the remaining 44,900 images as part of the unlabeled pool. MNIST dataset also consists test set of size 10,000 images and we used it to check the performance of our model after each stage of our active learning setup. We used budgets of 100 and 1000 samples for experiments \ref{fig:mnist}, resp. We used a similar setup for FashionMNIST. For CIFAR-10 which is similar to MNIST w.r.t total of no of images in train and test sets. CIFAR-10 training set out of which is sub-divided into 5000 images for the initial labeled pool, 5000 images  as a validation set, and the remaining images as part of the unlabeled pool, we used a budget of 2500 images per round of active learning, up to 40\% of the training data. CIFAR-10 test set consists of 10,000 images and we used it to check the performance of our model after each stage of our active learning setup} 

\textbf{Runs:} For all the experiments, we measured performance by computing the average accuracy across 5 independent runs.

\textbf{State of the art comparison}: We compared our method against several recent AL approaches including DAL-OSR\citep{mandivarapu2020deep}, Variational Adversarial Active Learning (VAAL) \citep{sinha2019variational}, Core-Set  \citep{sener2017active}, Monte-Carlo Dropout \citep{gal2016dropout}, Ensembles using Variation Ratios (Ensembles w. VarR ) \citep{freeman1965elementary} \citep{powerOfEnsemblesForActiveLearning}, Deep Bayesian AL (DBAL) \citep{gal2017deep}, \textcolor{black}{BatchBALD \citep{kirsch2019batchbald}, and WAAL(\citep{shui2020deep}).} As a baseline, we also included uniform random sampling (Random) since it remains a competitive strategy in the field of active learning.

\textbf{Architectures:} For experiments on MNIST and Fashion-MNIST we used a LeNET network \citep{LeNET} as the encoder, projector network, followed by a classifier. We used latent vectors of size 60. As noted in Sec.~\ref{sec:methodology}, the classifier consists of a single linear layer. For CIFAR-10, we used a VGG16 network \citep{simonyan2014very} as our encoder and a latent vector of size 512 followed by classifier with single layer.

\textbf{Optimization:} We optimized the overall system using a mini-batch size of 64, a learning rate of 0.001, barlow twins constant of 0.001  and a weight decay of $10^{-5}$. We optimized the system for 150 epochs at each stage and 20 epochs on MNIST. At the completion of a stage, using wiebull sampling method we requested labels for $b$ images from the unlabeled pool. Once the labels for the images are received from the oracle. These labeled images were added to the labeled pool and used in the subsequent learning stages.

\textbf{Image Augmentations}
We use the augmentations similar to BYOL \citep{grill2020bootstrap} which is used by major SSL approaches. As shown in the Fig \ref{fig:architecture} two distorted images are produced from given single  input image by applying different kind of transformations. The image augmentation pipeline starts with random cropping, resizing which was applied to all images. Followed by Gaussian blurring, color jittering, converting to grayscale,horizontal flipping, and solarization which were e last five are applied randomly,

\begin{figure}[h]
  \centering
  \includegraphics[width=8cm]{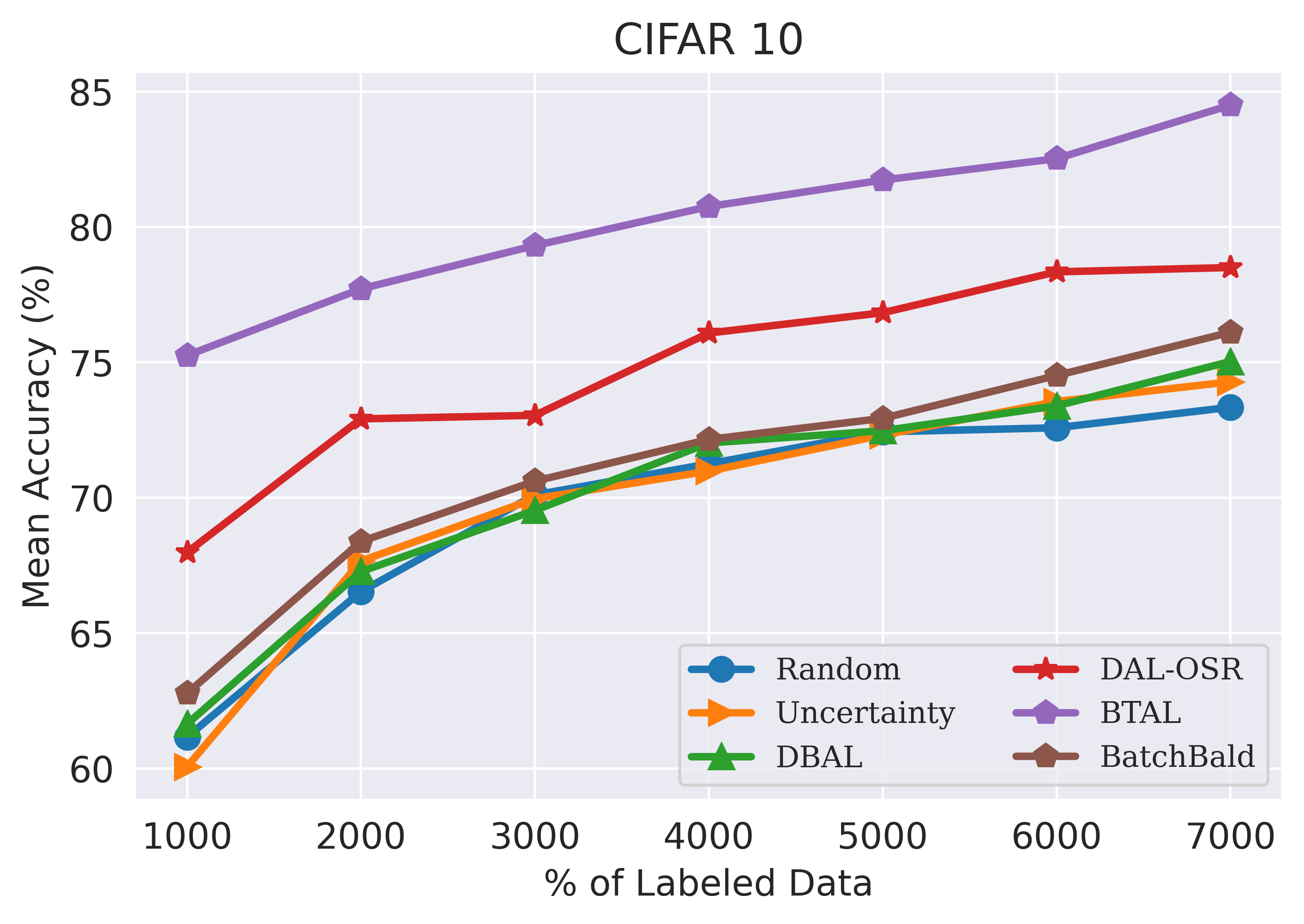}
  \caption{Robustness of our approach on CIFAR10 classification . Our results confirm that our approach significantly outperforms this baseline.}
  \label{fig:cifarr10_1000}
\end{figure}

\begin{figure}[h]
  \centering
  \includegraphics[width=8cm]{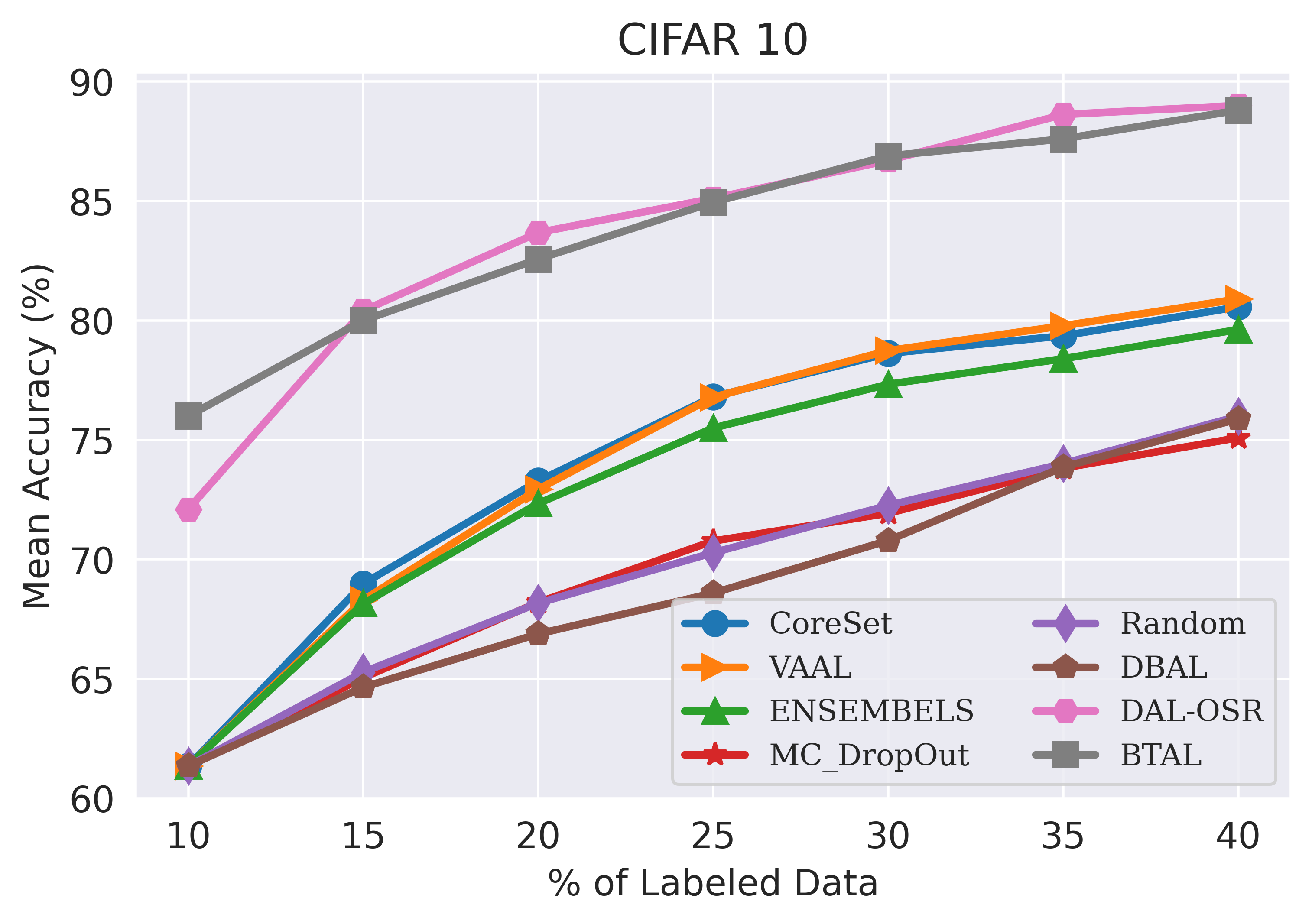}
  \caption{Robustness of our approach on CIFAR10 classification . Our results confirm that our approach significantly outperforms this baseline.}
  \label{fig:cifarr10_2500}
\end{figure}

\begin{figure}[h]
  \centering
  \includegraphics[width=8cm]{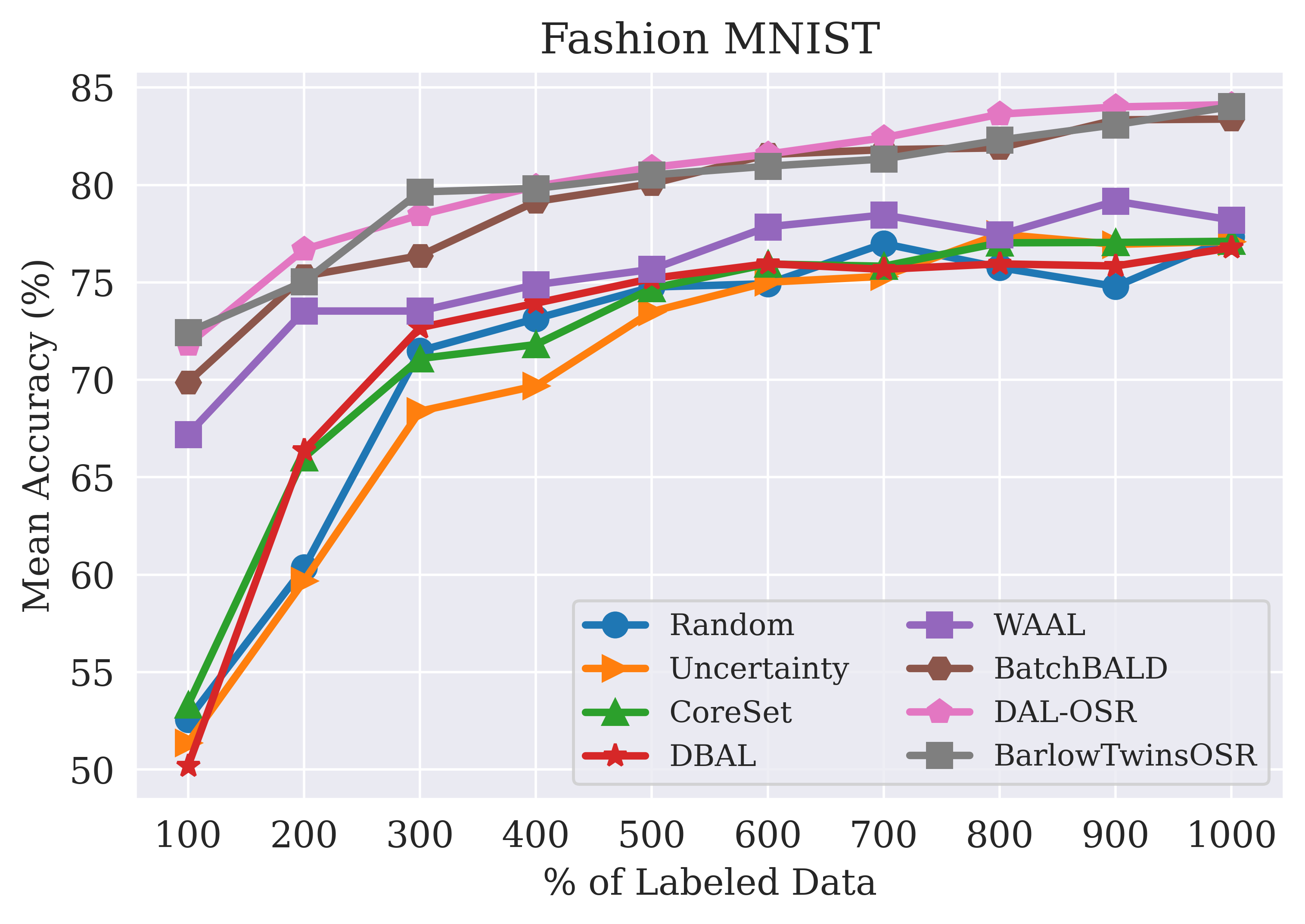}
  \caption{Robustness of our approach on CIFAR10 classification . Our results confirm that our approach significantly outperforms this baseline.}
  \label{fig:cifarr10_FashionMNSIT}
\end{figure}


\begin{figure}[h]
  \centering
  \includegraphics[width=8cm]{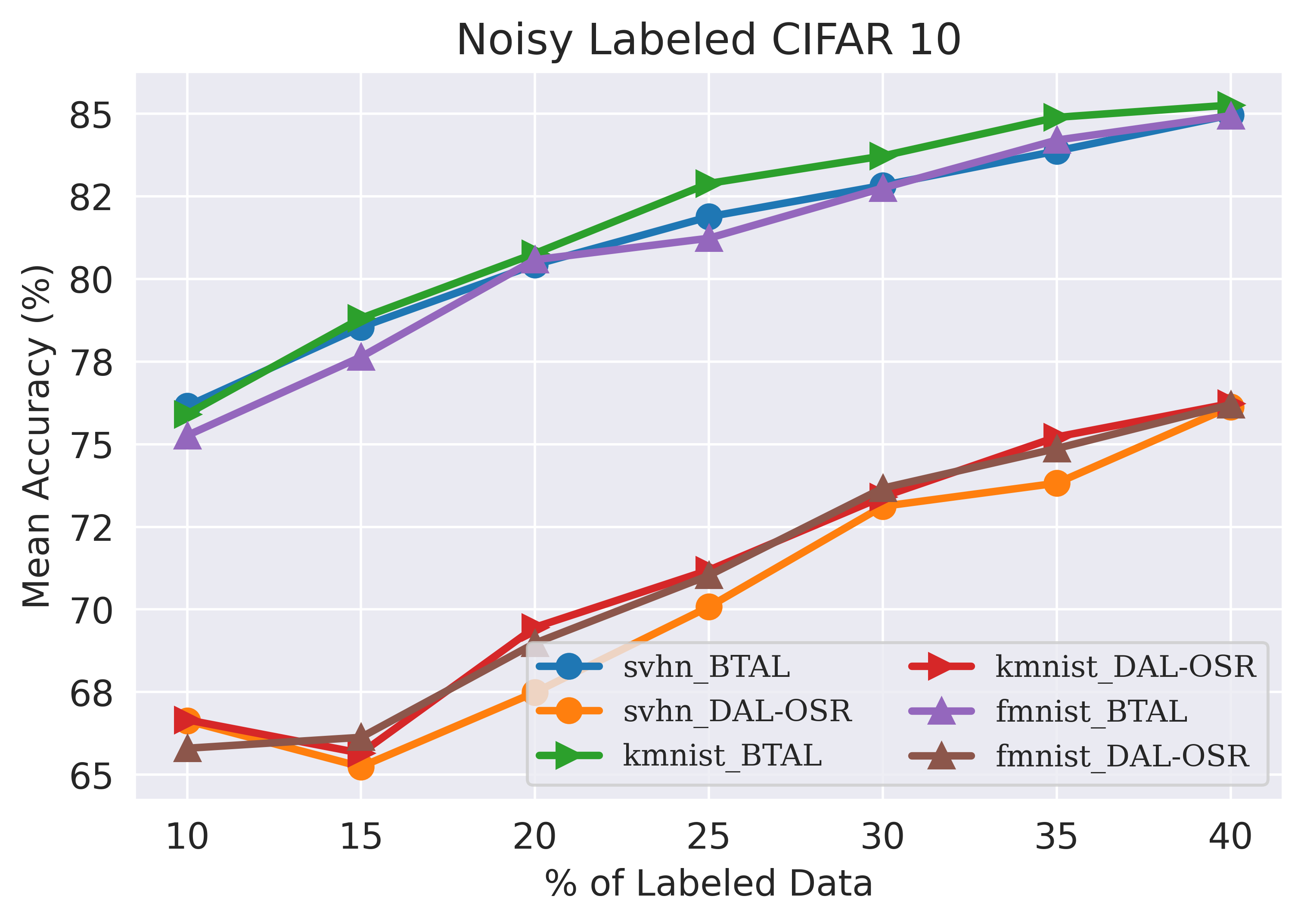}
  \caption{Robustness of our approach on CIFAR10 classification tasks when the unlabeled pool includes samples from either the SVHN, KMNIST, or FashionMNIST datasets. The first three curves used the $M_2$ classifier, while the ones with the 'Random' subscript used random sampling. Our results confirm that our approach significantly outperforms this baseline.}
  \label{fig:mixedUnlabeledPool}
\end{figure}

\textbf{Computer Vision Task results:}
To evaluate the effectiveness of our method we tested our method on MNIST,CIFAR-10, Fashion MNSIT and mixture of multiple datasets in the unlabeled pool.\\

\textbf{MNIST:} We conducted on MNIST dataset where size of initial labeled pool is 100 and using budget size of 100 at every stage of active learning. As it is shown in Fig \ref{fig:MNIST} our method performed on-par with the rest of exisitng method. As this is a easier computer vision task all the methods performed within the range.\\
\textbf{CIFAR-10:} We conducted two separate experiments for CIFAR-10 with different budget sizes. We conducted the experiment where the initial labeled pool is of size 5000 and budget($b$) is 2500 at each stage.  As shown in Fig. \ref{fig:cifarr10_2500} our proposed method performed on-par with the existing state of art method DAL-OSR and VAAL came in third, with an accuracy of only 80.71\% , followed by Core-Set with an accuracy of 80.37\%, and then Ensemble w VarR at 79.465\%. Random sampling, DBAL and MC-Dropout all trailed significantly behind other methods.\\

To evaluate the effectiveness of the proposed model when compared to other methods for small budgets we designed an experiments where the initial labeled pool is of size 5000 and budget($b$) is 1000 at each stage of active learning. As shown in the Fig. \ref{fig:cifarr10_1000} the proposed method overcomes all the existing methods by a huge margin and DAL-OSR method comes second followed by BAtchBALD and rest of the methods are in similar range. This proves the proposed method is very effective when the budget size if pretty low. The original accuracy which can be achieved using the entire CIFAR-10 dataset was 92.63\%.\\

\textbf{FashionMNIST:} To evaluate the robutness of our approach to different datasets we conducted experiemnt on another standard benchmark dataset FashionMNIST. Similar to previous experiments we compared our method with other exisiting state of the art methods like DAL-OSR, Core-Set  \citep{sener2017active}, Deep Bayesian AL (DBAL) \citep{gal2017deep},WAAL, \textcolor{black}{BatchBALD \citep{BatchBALD}, and WAAL(\citep{shui2020deep}).} As shown in the Fig \ref{fig:cifarr10_FashionMNSIT} our method outperforms all the methods and comes in-par performance with DAL-OSR.\\

\textbf{Mixed UnLabeled Pool:} Finally, we also tested the extreme case of active learning as proposed in DAL-OSR. We followed the similar setup follwed by DAL-OSR in which 10,000 images from other datasets like SVHN,KMNIST,KMNIST was mixed into source dataset of CIFAR-10. Thus the proposed method should distinguish not only between informative and non-informative samples
but also distinguish in-distribution data samples(CIFAR-10) from out-of-distribution samples(SVHN,KMNIST,KMNIST. The better model untilizes the budget well and picks the informative in-dataset samples. Eg: In the case of where the budget is 1000, If the model picks 1000 samples out of which 400 belongs to samples from out-of-distribution dataset. Then only 600 samples are sent for annonatation and add to labeled pool dataset and rest of the  400 samples added back to unlabled pool which makes total of 10,000 out of distribution data samples in the unlabeled pool at every state of active learning. As the active learning increses in stages it makes more and more difficult for the model to pick the in-label samples as unlabeled pool contains less in-label and more out of distribution samples.

\section{Conclusions and Future work}
\label{sec:conclusions}
In this work, we proposed a novel method for Active learning under iid and non-iid shift based unlabeled pool for computer vision based tasks using self-supervised learning technique along with wiebull sampling method. We evaluated our work by extensive comparisons with existing methods on three open source datasets. We rigorously benchmarked our method against the state-of-the-art active learning models on computer vision tasks. We also presented different budget based and mixed unlabeled pool setup studies to show the effectiveness of the proposed method with respect to the other methods. The results showed our method consistently performed on par and better than existing baselines on computer vision tasks. For future work, we would like to further explore more effective self supervised methods for handling active learning at scale.

\bibliography{reference}
\bibliographystyle{icml2021}

\end{document}